\documentclass[runningheads]{llncs}
\usepackage{graphicx}
\begin{document}
\title{Learning to Assist Agents by Observing Them}
%
%
\author{Antti Keurulainen\inst{1,3} \and Isak Westerlund \inst{3} \and Samuel Kaski\inst{1,2}
\and Alexander Ilin \inst{1}}
\authorrunning{Keurulainen et al.}
%
\institute{Helsinki Institute for Information Technology HIIT, Department of Computer Science, Aalto University
\and
Department of Computer Science, University of Manchester
\and
Bitville Oy, Espoo, Finland}
\maketitle              
\begin{abstract}
The ability of an AI agent to assist other agents, such as humans, is an important and challenging goal, which requires the assisting agent to reason about the behavior and infer the goals of the assisted agent. Training such an ability by using reinforcement learning usually requires large amounts of online training, which is difficult and costly. On the other hand, offline data about the behavior of the assisted agent might be available, but is non-trivial to take advantage of by methods such as offline reinforcement learning. We introduce methods where the capability to create a representation of the behavior is first pre-trained with offline data, after which only a small amount of interaction data is needed to learn an assisting policy. We test the setting in a gridworld where the helper agent has the capability to manipulate the environment of the assisted artificial agents, and introduce three different scenarios where the assistance considerably improves the performance of the assisted agents. 

\keywords{Deep reinforcement learning, cooperative AI, helper agent, meta-learning, modelling other agents.}
\end{abstract}

\section{Introduction}
The ability to build an AI system that can assist other AI agents and humans to reach their goals is an obviously important and ambitious goal. There are numerous possible applications that would benefit directly of such a capability. For example, helper agents could be harnessed to guide self-driving cars to allow smoother traffic flow. An example of cooperation with humans could be a helper agent, which is able to infer the goals of a human navigating a web site, and manipulate the website structure dynamically to help the human users reach their goals faster.

In this paper, we consider building such helper agents. We construct settings, where an artificial agent (\textit{a helper agent}) observes the behavior of an assisted agent (\textit{goal-driven agent}), and learns to perform useful assisting actions by manipulating the environment dynamics. 

We formulate the challenge as a meta-reinforcement learning problem. We first train a small population of goal-driven agents to reach their goals as fast as possible. The helper agent observes the goal-driven agents in many instances of the environments, and is trained by meta-reinforcement learning to perform useful assisting actions by receiving a positive reward when the goal-driven agents succeed in their tasks.

As a contribution of our work, we show that similar capabilities to inverse reinforcement learning arise spontaneously as a result of meta-learning. We demonstrate this effect and its usefulness for producing helping actions in three different scenarios:

\begin{enumerate}

\item By learning representations of the behavior of the goal-driven agent from the past partial trajectories in an unsupervised way. 

\item By learning representations of the behavior of the goal-driven agent from the past partial trajectories in a self-supervised way, by predicting the actions of the goal-driven agent.

\item By learning representations of the behavior of the goal-driven agent from the past partial trajectories in a supervised way, by predicting the goal of the goal-driven agent.

\end{enumerate}

\section{Related work}

There is a rich literature on reasoning about the behavior and inferring the goals of other agents \cite{Albrecht2018}, which is closely related to our work. Unlike most of the work related to modelling other agents, our specific goal was to build agents which can do assisting actions, based on the inferred representations of the behavior of the other agent.

Similar settings have been studied in multi-agent scenarios. As an example, in \cite{Dimitrakakis2017} a two-agent helper-AI scenario is presented, where both agents have the same reward function, but disagree on the transition dynamics of the Markov decision process (MDP). In our setting, in contrast, the action spaces of the agents are different, and the assisted agent is unaware of the assisting agent. The authors in \cite{Grover2018} consider learning representations of agent behaviors in multi-agent scenarios. Our approach is similar in that we also learn to represent agents' behaviors, but rather than communicating cooperatively, the helper agent attempts to change the environment in a beneficial manner.

An individual is considered to have a Theory of Mind (ToM), if they impute a mental state to themselves or others \cite{Leslie2004,Premack1978}. An inferred mental state can be used to infer goals and make predictions of behavior of other agents. Our work is inspired by Theory of Mind, as we are training behavioral embeddings of the goal-driven agents by only observing them. Examples of previous work implementing Theory of Mind in a machine learning context include \cite{Baker2017}, a Bayesian Theory of Mind model, and \cite{Rabinowitz2018}, which is based on deep learning. In our approach, a separate helper agent infers the goal-driven agent behavior and produces behavior embeddings as in \cite{Rabinowitz2018}, which are used for manipulating the environment in a beneficial way.

The problem of learning fast from limited amounts of data can be tackled with meta-learning methods, where the goal is to train a system in such a way that it is capable of performing fast adaptation to new situations \cite{Lemke2015,Schmidhuber1987}. We train the helper agent to be able to do few-shot learning of a new goal-driven agent in a new sampled environment by creating a behavior embedding of the agent based on small amounts of observations. Commonly used meta-learning methods include gradient based methods \cite{Finn2017} and memory-based methods \cite{Duan2016,Santoro2016,Wang2016}. Our meta-learning solution resembles mostly attention-based meta-learning \cite{Mishra2017}, since in our case the behavior embeddings are generated by averaging the state representations over the time steps without recurrent connections.

Our third scenario is partly inspired by the centralized training and decentralized execution (CTDE) concept \cite{Kraemer2016,Oliehoek2012}, where additional information can be used during the training but not during the execution. Instead of using private information of the goal-driven agent, we use information of the goal-driven agent's goal for pre-training the helper agent policy. 

Similar challenges could be solved by transfer learning or by using auxiliary tasks. In transfer learning, the goal is to transfer what has been learned in one task to some other final task, possibly by partially sharing parameters between the models \cite{Pan2010,Yosinski2014}. Numerous success cases have demonstrated benefits of transfer learning in the areas of computer vision \cite{Oquab2014} and natural language processing \cite{Alec2018,Devlin2019}. Our solution is one type of transfer learning, where the goal is specifically to pre-train one specific part of the policy function to allow faster learning for producing assisting actions. When using auxiliary tasks, the common representation among the original task and auxiliary tasks can be improved by adding extra auxiliary heads to the representation \cite{jaderberg2016reinforcement,mirowski2016learning,zhang2016augmenting}. In our solution, instead of adding an extra auxiliary head to the common representation, the helper agent is pre-trained on the task of making predictions of the goal-driven agents' actions in self-supervised manner or goals in a supervised manner. After completing the pre-training task, the final task training benefits from this ability. 

Inverse Reinforcement Learning (IRL) \cite{Ng2000} has similar goals, but does so by recovering the unknown reward function of the expert user or any agent it tries to model. The difference in our method is that we do not attempt to recover the reward function but rather model the agent behavior directly by observing the agent actions, and then build the helper policy, which uses the behavior embeddings. This approach is useful when moving to behavior where reward functions are complex or intractable, as often is the case when attempting to model human-like complex behavior.

\section{Setting}

We use a meta-reinforcement learning formulation for training both the goal-driven agents and the helper agent. The MDP is defined by a tuple $ M = \langle \mathcal{S}, \mathcal{A}, \mathcal{P}, \mathcal{R}, \gamma, \rho_0, H \rangle $, where $\mathcal{S}$ is the state space and $\mathcal{A}$ is the action space, $\gamma$ is the discount factor, $\rho_0 : \mathcal{S} \rightarrow [0,1] $ is the initial state distribution, $\mathcal{P} : \mathcal{S}  \times  \mathcal{A} \times \mathcal{S} \rightarrow [0,1]$ is the transition function, $\mathcal{R}: \mathcal{S} \times \mathcal{A} \rightarrow {\rm I\!R}$ is the reward function and $H$ is the horizon.

The agents maximize the expected sum of discounted rewards over the episodes $J(\pi_\theta) = E_{\tau\sim p_\pi (\tau)}[\sum_{t=0}^{H}\gamma^t \mathcal{R}(s_t,a_t)]$, where $\tau = (s_0,a_0, ...)$ is the trajectory of states $s_t$ and actions $a_t$ of the agent at time step $t$ for an episode of length $H$. 

The initial state is sampled from the distribution $\rho_0$ and the agent $j$ samples the action $a_t$ from its policy function $\pi_j(a_t|s_t)$. The next state is sampled from the transition dynamics function $s_{t+1} \sim \mathcal{P}(s_{t+1}|s_t,a_t)$. The agents learn through many episodes. For each episode, a new MDP $M_i$ is sampled from a family $\mathcal{M}$ of MDPs. The MDPs for the helper agent and the goal-driven agents share the same state space $\mathcal{S}$ but have different actions spaces and reward functions denoted as $\mathcal{A}^{helper}$ and $\mathcal{A}^{goal-driven}$, and  $\mathcal{R}^{helper}$ and $\mathcal{R}_j^{goal-driven}$. From the helper agent point of view, the behavior of the goal-driven agent is embedded in the transition dynamics of the sampled MDP, and each sampled MDP refers to different task in the meta-reinforcement learning framework. The meta-learning goal is to maximize the expected returns over the task distribution.

\section{Methods}

\subsection{The gridworld}
\begin{figure}[t]
\centering
\includegraphics[width=0.6\textwidth]{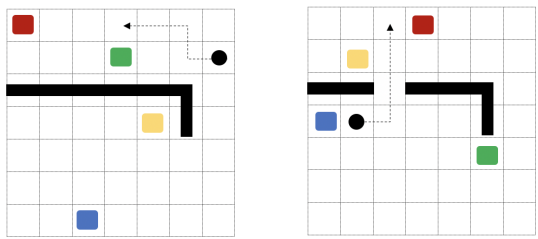}
\caption{Illustration of the 7x7 gridworld. An agent trajectory is illustrated with a dotted line, when the goal-driven agent is walking towards the red goal. Left: the helper agent has not opened a shortcut. Right: The helper agent has opened a shortcut} \label{gridworld}
\end{figure}

We test our architectures in various gridworld scenarios. Every new sampled MDP has a 7 x 7 gridworld environment, which includes four possible goals in random locations, an L-shaped wall in a random location and a goal-driven agent starting in a random position. Each goal-driven agent is trained to navigate to one of the goals with the shortest route. The wall has always the same basic shape, and its location is sampled but restricted so that the long edge of the wall needs to be at least two steps from the border of the gridworld. The left side of Figure \ref{gridworld} illustrates a sampled gridworld environment. 

The environment delivers a set of tensors as a state representation, one tensor per one time step. The state representation is a collection of 7x7x6 tensors, where the different channels in one tensor represent goal-driven agent location (1 channel), wall elements (1 channel) and goals (4 channels).

The episodes are divided in two distinct stages. In the first stage, the helper agent observes several partial trajectories of the unassisted goal-driven agent navigating to its goal in different sampled environments. In the second stage, a new environment is sampled and the same goal-driven agent is set in a random location. In the beginning of the second stage, the helper agent observes the state (a single tensor) and will decide on an action based on this observation. The right side of Figure \ref{gridworld} illustrates a situation where the helper agent has acted and opened a shortcut for the goal-driven agent.

\subsection{The goal-driven agents}
The goal-driven agents, which are part of the environment of the helper agent, are trained with the Proximal Policy Optimization (PPO) algorithm \cite{Schulman2017}. We train four different agent types which have different target goals. The action space $\mathcal{A}^{goal-driven}$ consists of five possible actions (move up, down, left, right and stay). The episode ends with a positive reward when the agent reaches the correct goal from four possible goals. Each step has a small negative reward. During the training of the goal-driven agent, there is a shortcut in the wall with the probability 0.5.

When the goal-driven agent $j$ follows its policy, it will produce a trajectory consisting of a list of state-action pairs
$$ \tau_{j} = \Big\{  \big( s_t, a_t \big)   \Big\}_{t=0}^H  $$
where  \(H\) is the length of the episode.

\subsection{The helper agent}

Once a population of goal-driven agents exists, the helper agent is trained with the PPO algorithm to infer the agent types by observing them, and to conduct an assisting action by manipulating the transition dynamics (creating a shortcut in a specified location). The action space $\mathcal{A}^{helper}$ of the helper agent consists of two actions, either to act or not to act, and the action is performed once at a specified time point depending on the scenario in question, thus the horizon $H$ for the helper agent is a single time step. The reward function $\mathcal{R}^{helper}$, is the same as with the goal-driven agent but augmented with a negative reward for making an action to assist the goal-driven agent. In other words, the helper agent gets a reward when the goal-driven agent succeeds in its task. The helper agent has access only to the partial trajectories of \(\tau_{j}\) of the goal-driven agents and thus not to their reward function \(R_j^{goal-driven}\) or policy  \(\pi_j^{goal-driven}\).

\section{Experiments and results}

\subsection{End-to-end training without agent-specific inference.}

We first evaluate the performance of the helper agent in a simplified situation where it does not have access to the past observations of unassisted behavior of the goal-driven agents. The results of this experiment serve as a lower bound for the helper agent performance. The helper agent does not observe the first stage of each episode and needs to decide on its actions only based on the initial state of the second stage. Thus, in this case the helper agent does not have the capability of inferring the behavior of the goal-driven agents.

In this scenario, the helper agent is a neural network with two convolutional layers and one fully-connected layer (see Figure \ref{scenario1}, left). More details of the neural network implementation is available in the appendix.

The green line in Figure \ref{main_results} illustrates the training result in this setting. The helper agent learns to improve its policy slightly despite not being able to infer the agent behavior. In this case, the helper agent tries to find the optimal performance by observing the general behavior of all agent types and by finding a more general policy that does not depend on a single agent's goal. As an example, if all goals are on the other side of the wall, it would be beneficial to open the wall. Likewise, if all goals are on the same side of the wall as the goal-driven agent, it is not useful to open the wall and take the penalty. These two scenarios are illustrated in Figure \ref{scenario1}. 

\begin{figure}[t]
\centering
\includegraphics[width=.7\textwidth]{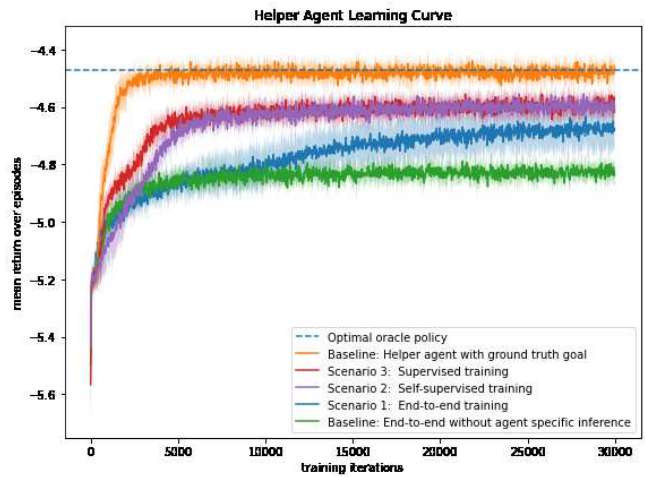}
\caption{The average learning curves of simulations with 8 random seeds with various helper agent architectures. The green line represents the learning curve without behavior embeddings and serves as a lower bound for the helper agent performance. The orange line shows the performance when the policy network uses ground truth information about the agent behavior, and serves as the upper bound for the helper agent performance. The blue line represents end-to-end training with embeddings (scenario 1), the purple line with action prediction pre-training (scenario 2) and the red line with goal prediction pre-training (scenario 3). The shaded region illustrates the standard deviation over 8 seeds.} \label{main_results}
\end{figure}

\begin{figure}[t]
\centering
\includegraphics[width=0.9\textwidth]{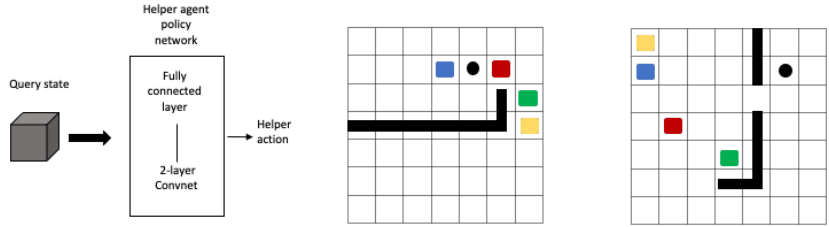}
\caption{Left: A simplified version of the helper agent that is not capable of inferring the goal-driven agent behavior. Middle: The helper agent has not opened the wall because all the goals are on the same side as the agent.
Right: The helper has opened the wall because all the goals are on the other side. The colored squares indicate the goals, black lines represent the wall and the black circle is the location of the goal-driven agent.} \label{scenario1}
\end{figure}

\subsection{Scenario 1: End-to-end training with agent-specific inference.}

\begin{figure}
\centering
\includegraphics[width=0.3\textwidth]{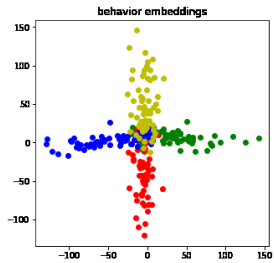}
\caption{Visualization of the 2-dimensional behavior embedding in scenario 1. The behavior inference part of the policy has spontaneously clustered the goal-driven agents based on their goals.}
\label{behavior embedding}
\end{figure}

\begin{figure}
\includegraphics[width=0.9\textwidth]{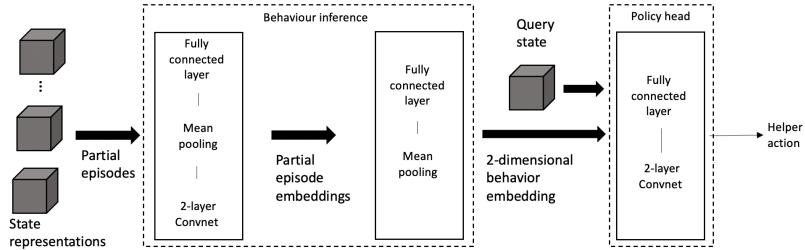}
\caption{Scenario 1. The helper agent architecture. The behavior inference part infers the goal driven agent behavior by observing partial unassisted behavior. The policy head produces the action distribution from a query state and the behavior embedding.}
\label{scenario2_1}
\end{figure}

In the first scenario, the helper agent is exposed to the past observations of unassisted behavior of the goal-driven agents. In the first stage, it now observes four first steps from four different episodes of the same agent but with different sampled layout of the gridworld. By observing these partial trajectories before conducting the assisting action, the helper agent has a chance to learn about the goal-driven agents' goals based on the agents' unassisted behavior. The helper agent architecture is illustrated in Figure~\ref{scenario2_1}. During the first stage, the helper agent uses the \textit{behavior inference} part of the policy. The embeddings of partial episodes are formed by using a convolutional network, followed by mean pooling and a fully connected layer. The behavior embeddings are formed in a similar manner from the episode embeddings but without convolutional layers. During the training, the behavior inference part learns to construct 2-dimensional behavior embeddings by clustering the agents based on their goals, where each cluster contains agents with the same goal (Figure~\ref{behavior embedding}).

The \textit{policy head} part of the helper agent has the same structure as the simpler case shown in the left side of Figure~\ref{scenario1}. The task of the policy head is to produce the action distribution but now it additionally receives the behavior embedding from the behavior inference part. The training of this scenario is conducted end-to-end.

The result of this experiment is illustrated by the blue line in Figure \ref{main_results}. The helper agent is capable to use the information produced by the behavior inference part and as a result the performance is improved considerably. Since there is significant variance present in simulations with different seeds, we can assume that the end-to-end training is more sensitive to the initialization than the other scenarios.

\subsection{Scenario 2: Self-supervised pre-training with agent-specific inference.}

In this scenario the helper agent has an initial pre-training phase where it observes multiple partial episodes where the goal-driven agents behave without assistance, simulating a situation where previously collected offline data of the assisted agent behavior is available. The pre-training is conducted in a self-supervised fashion by predicting the action distributions of the goal-driven agents. After the partial rollouts of episodes the distribution over the next action is predicted by using a linear classification layer on top of the 2-dimensional behavior embedding. The cross-entropy loss is calculated between the observed one-hot action and the predicted action distribution. Once the pre-training is complete, the linear classification layer is discarded and the setting is converted to a similar one as in scenario 1. 

Pre-training of the behavior inference part with action predictions clearly helps the policy function to learn assisting actions as the learning is faster and there is less variation across different seeds, as is visible by the purple line in Figure \ref{main_results}.

\subsection{Scenario 3: Supervised pre-training with agent-specific inference.}

In our final scenario the helper agent has again an initial pre-training phase where it observes multiple partial episodes where goal-driven agents behave without assistance. Inspired by the centralized training decentralized execution (CTDE) method \cite{Oliehoek2012,Kraemer2016}, the helper agent is now provided with the information about the agent’s goal in each episode during the pre-training phase, but not during the execution after the pre-training. The pre-training is conducted in a similar manner as in scenario 2 by using the linear classifier on top of the 2-dimensional behavior embedding, and by calculating the cross-entropy loss between the one-hot goal label and the predicted goal distribution. In this case, the pre-training has strong supervision, and the helper agent is able to cluster the goal-driven agents based on their goals already during the pre-training phase. This allows faster learning as is visible with the red line in Figure \ref{main_results}.

\subsection{Summary of results}

In order to evaluate the performance of the helper agent in various scenarios, we made additional measurements that are useful for gaining insight of the helper agent performance. We measured the mean return over episodes when the wall was permanently closed, when it was permanently open and the optimal policy by making rollouts from the query state. For each query state, we made a rollout both with wall open and wall closed scenarios. By selecting higher return from these rollouts, we measured the optimal policy. All key results are summarized in Table \ref{tab1}. The results show, that even when the helper agent does not have the possibility to observe the behavior of the goal-driven agent, the performance is improved. The performance is further increased when the helper agent can observe partial episodes of the goal driven agent (scenario 1). The self-supervised pre-training (scenario 2) and supervised pre-training (scenario 3) achieve highest overall performance as they are closest to the optimal performance. Interestingly, the self-supervised scenario reach a similar performance as the supervised pre-training, which suggests that self-supervised method is a potential alternative for assessing behavior when labels are not available for the pre-training.

\begin{table}
\caption{Helper agents with progressively more capabilities are able to help the goal-driven agents more.}\label{tab1}
\begin{tabular}{|l|l|l|}
\hline
Model &  Mean return \\
\hline
Baseline: Wall always open (with penalty for opening the wall) &   -5.19 \\
Baseline: Wall always closed &  -5.85 \\
Baseline: Helper agent with ground truth goal &  -4.48  $\pm$0.03 \\
(oracle) optimal policy & -4.47 \\
Baseline: End-to-end without agent specific inference & -4.82 $\pm$0.03 \\
Scenario 1: End-to-end with agent-specific inference & -4.66 $\pm $0.08 \\
Scenario 2: Self-supervised pre-training with action predictions & -4.59 $\pm$0.04 \\
Scenario 3: Supervised pre-training with goal predictions & -4.59 $\pm$0.03 \\
\hline
\end{tabular}
\end{table}

\section{Discussion}

In this paper, we introduced several scenarios, where a helper agent can assist goal-driven agents by manipulating the transition dynamics of the environment. We showed that by using meta-learning, similar capabilities to inverse reinforcement learning arise spontaneously. Our method is also an example how previously collected data can be used to enhance the performance of a helper agent when offline learning methods are not possible. This kind of situation arises if the previously collected data is from the assisted agent behavior, not from a helper agent behavior.

The results indicated that by learning to infer the agent's behavior, the helper agent's policy can be significantly improved. Furthermore, the training can be made faster and more robust by a separate pre-training phase. The pre-training can be done either in self-supervised manner by predicting the actions of the assisted agent, or by using additional information about the agents' goals that is available during the training but not during the execution. This kind of scenario is valid when it is of interest to predict a known feature that explains the behavior, such as the goal, skill level, level of stochasticity, level of visibility, or level of memory. An interesting continuation to this work could be extending to modeling more complex human-like behaviors in scenarios that are closer to real-world human-computer interactions (for example, web-site navigation).

\section{Acknowledgements}

This work was supported by the Academy of Finland (Flagship programme: Finnish Center for Artificial Intelligence FCAI, and grants 319264, 292334).

%
%
%
%

\bibliography{helper.bib}

\appendix
\section{Neural Network details}

\subsection{helper agent}

The helper agent behavior inference part has two convolutional layers, batch normalization layers, and two fully connected layers. Convolutional layers 1 and 2: 8 filters, 3x3 kernels, stride 1, zero padding to maintain the dimensions. Followed by ReLU activations and batch normalization. The output of the convolutional layers of each episode is averaged across the time dimension to produce the representations of the partial episodes. The representations are fed through a linear layer, averaged across the episodes of the agent and fed through another fully connected layer to produce the 2-dimensional behavior embeddings.

The policy head part of the helper agent has a shared policy and value head backbone with 2 convolutional layers with 16 filters, 3x3 kernels, stride 1 and zero padding. The output of the convolution is followed by a policy head with Softmax to produce action probabilities and a separate value head to produce value estimates.

The helper agent is trained using the Adam optimizer with a learning rate of 1e-4. Reward for opening the wall: -1. Reward of every step: -1. Reward of reaching the goal +1.

Using a batch size of 256 simulations were run for 60 000 PPO training steps and pre-training was done using 10 000 classification steps. 
The baseline results where the wall is always closed, the wall is always open and the optimal policy were calculated as the mean return over 512 000 rollouts. The helper agent results were calculated as the mean of the last 51 200 rollouts during training.

\subsection{Goal-driven agents}

The goal-driven agent policy network has two convolutional layers and a fully connected layer. 

Convolutional layer 1: 16 filters, 3x3 kernels, stride 1, ReLU activations, zero padding to maintain the dimensions. Convolutional layer 2: 32 filters, 3x3 kernels, stride 1, ReLU activations, zero padding. Fully connected layer with 32 hidden units: ReLU activation followed by a policy head with Softmax to produce action probabilities and a value head to produce value estimates.

Reward of every step: -1. Reward of reaching the goal +1.

\end{document}